\documentclass[conference]{IEEEtran}
\IEEEoverridecommandlockouts
\usepackage{cite}
\usepackage{amsmath,amssymb,amsfonts}
\usepackage{algorithmic}
\usepackage{graphicx}
\usepackage{graphics}
\usepackage{hyperref}
\usepackage{array}
\usepackage{lipsum}
\usepackage{float}
\usepackage{multirow}
\usepackage{textcomp}
\usepackage{xcolor}
\def\BibTeX{{\rm B\kern-.05em{\sc i\kern-.025em b}\kern-.08em
    T\kern-.1667em\lower.7ex\hbox{E}\kern-.125emX}}
\begin{document}

\title{Which Emoji Talks Best for My Picture?
}

\author{\IEEEauthorblockN{Anurag Illendula}
\IEEEauthorblockA{\textit{Department Of Mathematics} \\[0em]
\textit{IIT Kharagpur}\\[0em]
Kharagpur, India \\[0em]
aianurag09@iitkgp.ac.in}
\and
\IEEEauthorblockN{Kv Manohar}
\IEEEauthorblockA{\textit{Department Of Mathematics} \\[0em]
\textit{IIT Kharagpur}\\[0em]
Kharagpur, India \\[0em]
kvmanohar22@iitkgp.ac.in}
\and
\IEEEauthorblockN{Manish Reddy Yedulla}
\IEEEauthorblockA{\textit{Department of Engineering Science} \\[0em]
\textit{IIT Hyderabad}\\[0em]
Hyderabad, India \\[0em]
es15btech11012@iith.ac.in}
}

\maketitle

\begin{abstract}

Emojis have evolved as complementary sources for expressing emotion in social-media platforms where posts are mostly composed of texts and images. In order to increase the expressiveness of the social media posts, users associate relevant emojis with their posts. Incorporating domain knowledge has improved machine understanding of text. In this paper, we investigate whether domain knowledge for emoji can improve the accuracy of emoji recommendation task in case of multimedia posts composed of image and text. Our emoji recommendation can suggest accurate emojis by exploiting both visual and textual content from social media posts as well as domain knowledge from Emojinet. Experimental results using pre-trained image classifiers and pre-trained word embedding models on Twitter dataset show that our results outperform the current state-of-the-art by 9.6\%. We also present a user study evaluation of our recommendation system on a set of images chosen from MSCOCO dataset.

\end{abstract}


\section{Introduction}

\begin{figure}
\label{sec:instagramstatistics}
\centering
\includegraphics[width=1.0\linewidth]{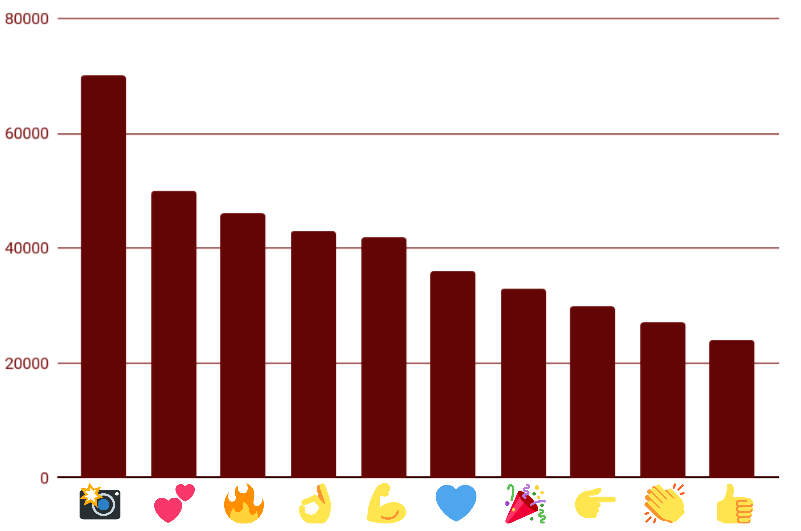} 
\caption{Usage of most frequently occurring emojis on Instagram. The Statistics are extracted from \url{https://www.quintly.com/blog/2017/01/instagram-emoji-study-higher-interactions}}.
\label{fig1}
\end{figure}

The word emoji has originated from the Japanese language with the letter ``e'' (meaning picture) and ``moji'' (meaning character). Emojis are considered to be the $21^{st}$-century transformation of emoticons. They were initially developed in 1999 as a 12*12 pixel grid by Shigetaka Kurita as a part of a Japanese team working on an early version of a mobile internet platform. But present emoji images can be scaled to unlimited resolution with the help of vector graphics\footnote{\url{https://cnn.it/2MitfM2}}. The utilization of emojis in social media has visually perceived a rapid increase over the last few years as they have become a way to integrate a tone and non-verbal context to daily communication.  According to the latest statistics released by Twitter, tweets with images links get 2x the engagement rate of those without\footnote{\url{https://bit.ly/1u31GbO}}. The study of emoji usage in social media platforms has been one of the exciting research fields, Instagram has reported that the emoji usage on their website has seen an increase of 10\% in a single month after the release of Emoji keyboard on iOS mobiles in 2011, and it also reported that more than 50\% of all captions and comments include an emoji or two\footnote{\url{https://engt.co/2JFJlxz}}. Hence considering the extensive usage of images on their platform, Instagram has started analyzing profiles of users who use emojis, and they reported that 53\% of users use emojis in their posts\footnote{\url{https://bit.ly/2EnbxSE}}. \hyperref[sec:instagramstatistics]{Figure 1} shows the usage of different emojis in Instagram. Images being more expressive than text, sending a message utilizing one diminutive emoji is more efficacious than a text message is the other important feature which increased emoji usage in social media. According to the latest statistics released by Emojipedia, there are 2666 emojis which are further divided into different subcategories. Earlier in 2016 most search engines namely Google and Bing supported emoji search\footnote{\url{https://selnd.com/2t4vjyk}}, but in 2017 Twitter has also enabled users to search for tweets using emoji as a keyword\footnote{\url{https://bit.ly/2sUVWGy}}.

Due to limited linguistic concepts, earlier manually annotated patterns were used as external knowledge concepts to enhance NLP systems. With the advent of advanced knowledge base construction, large amounts of semantic and syntactic information are made available which helped researchers enhance the performance of most NLP systems namely word embedding models\cite{bian2014knowledge} and many other prediction and classification tasks \cite{yang2017leveraging,xing2017topic}. In recent years several noteworthy large, cross-domain knowledge graphs have been developed. Researchers \cite{shin2015incremental} have also worked on incrementally populating knowledge graph from unstructured data which encompasses problems of extraction, cleaning, and integration. This research in the development of advanced knowledge graphs has enabled many researchers \cite{bian2014knowledge,yang2017leveraging,DBLP:conf/webi/ShethPWT17} to leverage external domain knowledge in Natural Language Processing (NLP) systems to improve machine understanding. External knowledge has also been effective to improve the accuracies of emoji understanding tasks including but not limited to emoji similarity \cite{wijeratne2017semantics}, emoji sense disambiguation \cite{emojineticwsm}. In this paper, we investigate whether external knowledge from EmojiNet can enhance the performance of emoji recommendation task in the context of images.
\vspace*{-3mm}
\begin{figure}[H]
\label{sec:example}
\centering
\includegraphics[width=1.0\linewidth]{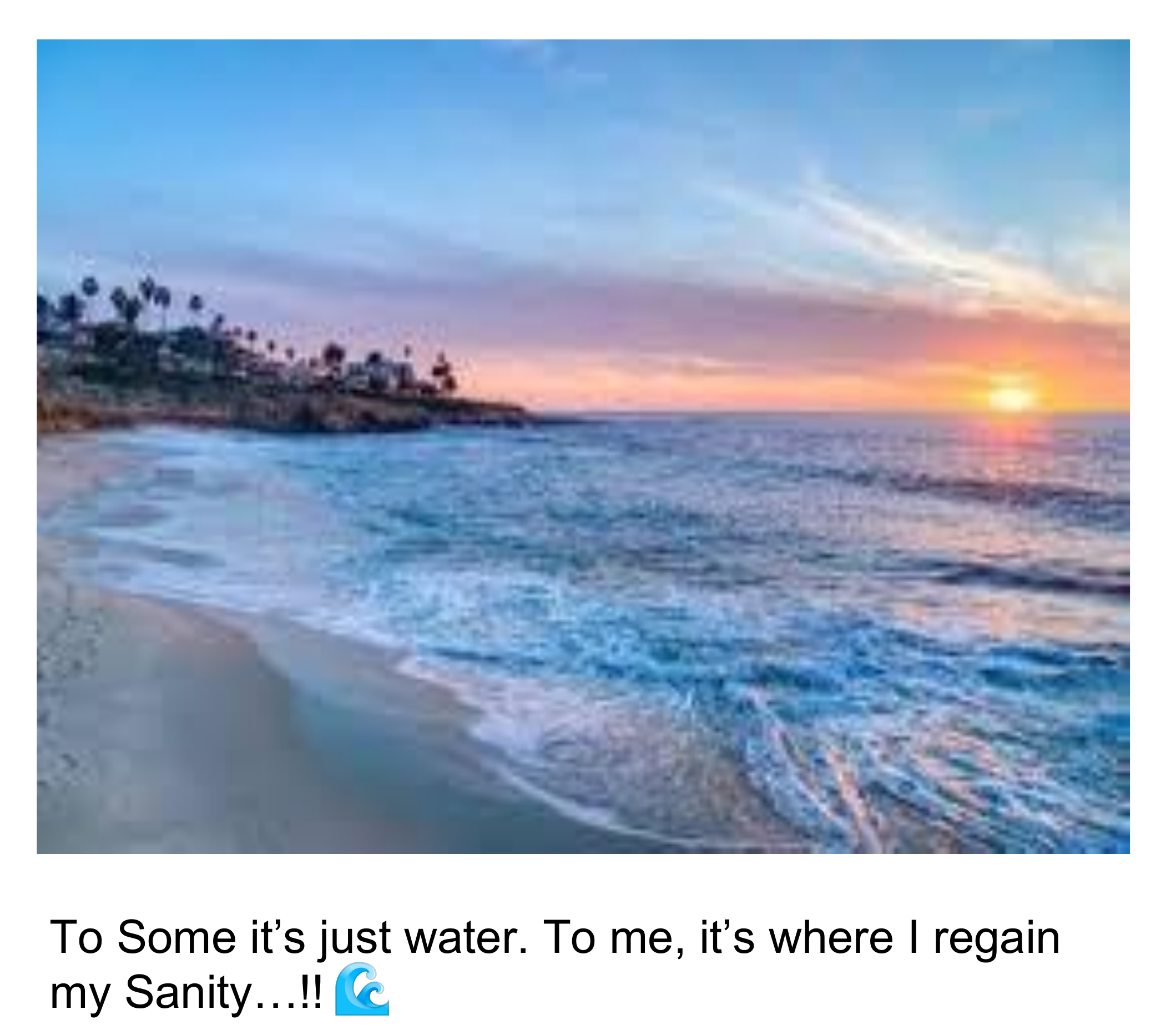} 
\vspace*{-7mm}
\caption{The emoji in the description is used to symbolize a "WATER WAVE" at the sea}.
\label{fig2}
\end{figure}
\vspace*{-3mm}

Image Classification is one of the fundamental challenges in the field of computer vision. There has been significant progress in the field of Computer Vision with the emergence of Convolutional Neural Networks (CNNs) \cite{KrizhevskySH17}. These Deep Convolutional Neural networks have led to series of breakthroughs for various image processing tasks including but not limited to image classification \cite{KrizhevskySH17}, object detection \cite{GirshickDDM13}, and semantic segmentation \cite{ShelhamerLD16}. Deep CNN's integrate the low/ high-level features and classifies in end-to-end multi layer fashion, and the ``levels" of features can be enriched by increasing the number of stacked layers \cite{zeiler2014visualizing}. All the current state-of-the-art techniques for Computer vision and Natural language processing tasks rely heavily on labelled data. The current state-of-the-art in image classification includes Deep Residual Networks \cite{He2015} which consists of shortcut connections between the stacked layers of the Deep CNN and residual representations. As our task for emoji recommendation requires us to classify the image effectively, we use the Deep Residual Neural networks which is the state-of-the-art image classifier to achieve better results for emoji recommendation.

With the rapid growth of emoji, not only they are being used with text, but also being used in the context of images to provide additional contextual clues on what is depicted in an image. Consider the image shown in \hyperref[sec:example]{Figure 2}. The user who posted this tends to use an emoji which relates to one or more of entities at that can be used to describe seashore. In this example we see the emoji, ``water wave'' \includegraphics[height=1em]{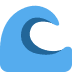} which is used to symbolize a water wave at sea. We hypothesize that having access to the images posted on social media can help recommend an emoji that can be used in the description of the image.

In this paper, we present an approach which combines visual concepts, user descriptions and external knowledge concepts from EmojiNet (the only machine-readable inventory which enables computers to understand emojis) to predict an emoji in the context of an image. We evaluate our approach on Twitter dataset crawled using Twitter API, and we also evaluate our approach using a set of manually annotated images from MSCOCO dataset. We plan to release the complete dataset upon the acceptance of the paper to help other researchers. Most of the current natural language processing and computer vision tasks involving text or image processing rely on manually annotated data. To create a gold-standard dataset to evaluate our approach, we asked human annotators who are knowledgeable with the usage of emojis to select an emoji from the complete set of emojis which they intend to use in the context of the image and description. We label each image with an emoji that is selected most number of times by the annotators. \hyperref[sec:dataset]{Section 3} further explains the creation of our evaluation datasets. We also compare our accuracy with the previous state-of-the-art approaches for emoji prediction in the context of images; experimental results show that our model outperform the previous state-of-the-art Image2Emoji models developed by Cappallo et al. and Barbieri et al.  \cite{cappallo2015image2emoji,barbieri2018multimodal}

 
In the rest of this paper, we first discuss the related work done by other researchers in \hyperref[sec:related]{Section 2}. In \hyperref[sec:dataset]{Section 3}, we discuss the creation of evaluation datasets and discuss our model architecture in \hyperref[sec:model]{Section 4}. In \hyperref[sec:experiments]{Section 5} we conduct extensive experiments to evaluate our method of approach. Finally, we discuss the observed results using our approach and conclude in \hyperref[sec:discussion]{Sections 6} and \hyperref[sec:conclusion]{Section 7}, respectively. The source code and annotated dataset will be made available upon the acceptance of the paper to help other researchers.


\section{Related Work}
\label{sec:related}

Content-Based Information Retrieval (CBIR) is the historic line of research in multimedia. This task usually deals with retrieving images in the dataset that are most similar to the query image. Bag-of-Words representations \cite{zhang2010understanding} has seen a sustained line of research in this task which has been effective up to million sized image datasets. This representation first consists of describing an image with a set of local descriptors such as Scale Invariant Feature Transform (SIFT) \cite{lowe2004distinctive}, and then in aggregating these descriptors into a single vector that collects the overall statistics of so-called ``visual words". Recently another step towards CBIR has achieved by VLAD \cite{li2018multiple}. This Information Retrieval task has eventually led to research in image classification which is one of the fundamental challenges in Computer Vision. Convolutional Neural Networks (CNNs) \cite{KrizhevskySH17} have shown promising results in image classification. Research on shortcut connections has been an emerging topic since the development of multi layer perceptrons which has shown promising results for Image processing tasks. Generally, these multiple layers have been connected using shortcut connections using gated functions \cite{hochreiter1997long}. In image classification, depth of the network, i.e., number of layers within the network is of crucial importance as noted by Simonyan et al. \cite{Simonyan14c}. Increasing the depth can have an adverse effect for image classification task. Most notably, vanishing/exploding gradients problem~\cite{pmlr-v9-glorot10a} and the degradation problem~\cite{He2015} are of significant importance. These problems are overcome by the introduction of shortcut connections and residual representations introduced in Residual Networks \cite{He2016} which won the 1st place in ImageNet 2015 Image classification competition \footnote{\url{https://bit.ly/2y4J8Cz}}. Residual Networks and its extensions which consist of many residual units have shown to achieve state-of-the-art accuracy for image classification tasks on datasets such as ImageNet \cite{ILSVRCanalysis_ICCV2013}.

Prior work on emoji prediction in text analytics has been done by Francesco et al. \cite{barbieri2017emojis, barbieri2018multimodal} and emoji prediction in case of images has been done by Cappallo et al. \cite{cappallo2015image2emoji,cappallo2018new}. Francesco et al. \cite{barbieri2017emojis} have worked on building models for emoji prediction in case of text messages especially twitter using state-of-the-art NLP techniques and also emoji prediction in case of images where they combined both visual and textual features for emoji prediction \cite{barbieri2018multimodal}. Their results have proved that visual features can help the model predict emoji accurately in multimedia datasets. Cappallo et al. worked on building an emoji recommendation system in the context of an image considering emoji names as external knowledge concepts for emoji prediction. This recommendation system relies on state-of-the-art image classification model to classify images and word embedding model to represent a word in a low dimensional vector space.

The idea of Semantic Web is that of publishing and querying knowledge on the Web in a semantically structured approach; it has been introduced to the wider audience by Berners-Lee \cite{berners2001semantic}. According to his vision, the web of documents must be extended such that the relationships between entities can be represented. This vision of Berners-Lee has led to the development of many structured knowledge bases where different entities are linked by relationships. Wordnet \cite{miller1995wordnet}, Freebase \cite{bollacker2008freebase}, and YAGO \cite{suchanek2007yago} are some of the manually constructed knowledge bases which deal with textual knowledge. Berners-Lee's vision has eventually led to the development of semi-automatically or automatically constructed knowledge graphs like DBPedia \cite{auer2007dbpedia} and NELL \cite{mitchell2018never}. Many attempts have been made by several researchers to embed the symbolic representations into continuous space which helps in statistical learning approaches \cite{bordes2011learning}. The extensive use of emojis in social media has been identified earlier, and this led to the development of the first emoji sense inventory EmojiNet by Wijeratne et al. \cite{emojineticwsm}.

Recent past has seen a rapid increase in the number of researchers working on using external domain knowledge to improve the accuracies of many NLP and Image processing tasks \cite{DBLP:conf/webi/ShethPWT17,yang2017leveraging,bian2014knowledge}. The reason being external knowledge helps the machine to understand the topics which can further aid in machine understanding. EmojiNet, the most extensive emoji sense inventory developed by Wijeratne et al. \cite{emojineticwsm} made vast amounts of linguistic knowledge available ranging from emoji sense labels to emoji sense definitions (textual descriptions which explain the context of use of different emojis). Recent research has shown that EmojiNet improved the accuracies of emoji similarity \cite{wijeratne2017semantics} and emoji sense disambiguation tasks \cite{emojineticwsm}. In this paper, we leverage external knowledge concepts from  EmojiNet to enhance the accuracy of emoji prediction task in the context of images.

Embeddings capture the semantics of a word and the syntactic information of the usage of the word in different contexts. Earlier many researchers have worked on building word embedding models to visualize words in low dimensional vector space. Earlier word2vec \cite{mikolov2013distributed} or GloVe \cite{pennington2014glove} have been the most popular word embedding models. But FastText word embedding model \cite{bojanowski2016enriching} has been even more effective in social NLP systems as the fastText model could learn sub word information. Many natural language processing tasks rely on learning word representations in a finite-dimensional vector space. Barbieri at al. \cite{barbieri2017emojis} and  Augenstein et al. \cite{eisner2016emoji2vec} have done prior work on learning emoji representations using the traditional approaches (CBOW and skip-gram models). Recent research showed that semantic embeddings are more efficient than the embeddings learned using the traditional approach as they inherit semantic and syntactic knowledge, and semantic embeddings have shown great success in similarity and analogical reasoning tasks \cite{bian2014knowledge}. Wijeratne et al. \cite{wijeratne2017semantics} have worked on learning semantic representations of emojis using knowledge concepts from EmojiNet, and these embeddings have improved the results of emoji similarity and other natural language processing tasks \cite{kursuncu2018s}. Recent research by Seyednezhad et al. \cite{seyednezhad2017understanding} and Fede et al. \cite{fede2017representing} has shown that emoji co-occurrence is one of the important features which helps us to understand the context of use of multiple emojis. Illendula et al. \cite{illendula2018learning} have worked on learning emoji representations using emoji co-occurrence network graph and state-of-the-art network embedding model, these embeddings out-performed the previous state-of-the-art accuracies for sentiment analysis task.

In this paper, we present an emoji recommendation system which effectively recommends an emoji in the context of an image. We use Res-Net \cite{He2015} image classifier which is the state-of-the-art image classifier, word representations trained on a corpus of descriptions of images in MSCOCO \cite{LinMBHPRDZ14} using Google-News pre-trained word embeddings and FastText \cite{joulin2016FastText} model which could learn sub-word information. We use the bag of words model developed by Wijeratne et al. \cite{wijeratne2017semantics} to learn the emoji embeddings which are used as external knowledge concepts. We report the results observed considering different emoji knowledge concepts from EmojiNet namely emoji names, emoji senses, emoji sense definitions and two different word embedding models in \hyperref[sec:experiments]{Section 5}. We also present the results obtained by our user study on MSCOCO dataset in \hyperref[sec:experiments]{Section 5}.


\section{Dataset Creation}
\label{sec:dataset}
\subsection{Twitter Dataset}

\begin{figure}[H]
\label{sec:emojicount}
\centering
\includegraphics[width=1.0\linewidth]{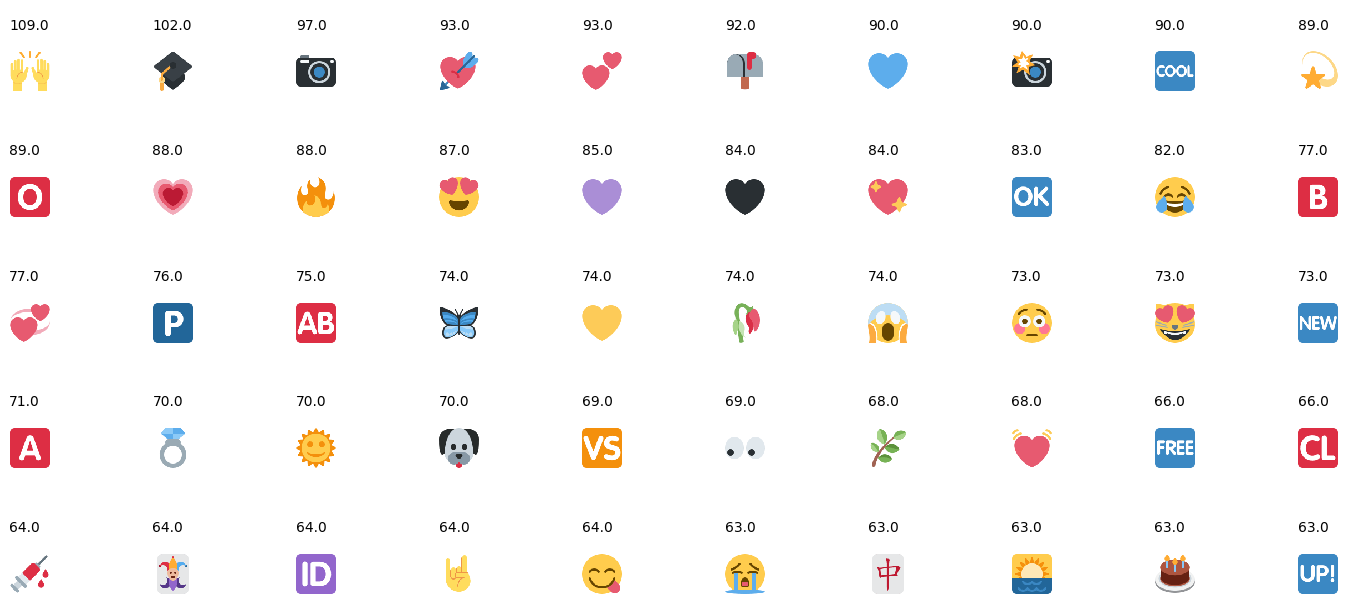} 
\caption{Distribution of Number of Images in Twitter Dataset}.
\label{fig3}
\end{figure}

We extracted tweets using the Twitter streaming API geo-localized in the United States of America considering each emoji as a keyword at a time for search from the list of 2389 emojis listed in EmojiNet\footnote{\url{https://bit.ly/2JDX0F0}}. We then filtered the tweets by considering only the tweets which are embedded with an image, and further filtered the dataset by considering the tweets which have only one emoji embedded in the tweet since our model couldn't learn the context of use of multiple emojis at the same time. During the process of filtration, we also ensured that the tweet has a textual description. We could extract 27136 tweets which have one of 1079 emojis, the distribution of the number of tweets of the dataset is shown in \hyperref[sec:emojicount]{Figure 3}. We consider the emoji embedded in the description as the label for the tweet, and we use our model to get a set of emoji recommendations in the context of the image with textual description. We then evaluate our emoji recommendation model and report the results in \hyperref[sec:experiments]{Section 5}.

\vspace*{-3mm}

\subsection{User Annotation}

We also evaluate our model on a set of 600 images from MSCOCO 2017 validation dataset\footnote{\url{https://bit.ly/2JHIhZX}} \cite{lin2014microsoft} which belong to different classes\footnote{\url{https://bit.ly/2mYUfDd}} listed in ImageNet Image Classification competition. We ensured that our evaluation dataset does not include multiple images of the same category as this would lead to biased results, and this filtration also allows us to verify the accuracy of our approach on different classes of images. These set of images in MSCOCO dataset are associated with a set five descriptions which explain the context of the image. We asked three annotators who are knowledgeable with the context of use of emojis to manually annotate the image with the textual description with an emoji from the complete set of 2389 emojis listed in EmojiNet. The human annotators are undergraduate students, two annotators from Indian Institute of Technology Kharagpur, and the other belongs to Indian Institute of Technology Hyderabad. The two annotators are aged between 18-23; one males and one female. The annotators were shown an image with the complete set of descriptions and asked to select an emoji which they wish to use to increase their expressiveness. Each image in our evaluation dataset is annotated by all the three annotators, and we assume that the emoji selected most times by the annotators as the emoji predicted in the context of the image. We use this annotated dataset to evaluate our model for emoji recommendation and report our results in \hyperref[sec:experiments]{Section 5}.


\section{Model}
\label{sec:model}
\subsection{Pre Training}

\begin{figure}[H]
\label{sec:bagof}
\centering
\includegraphics[width=1.0\linewidth]{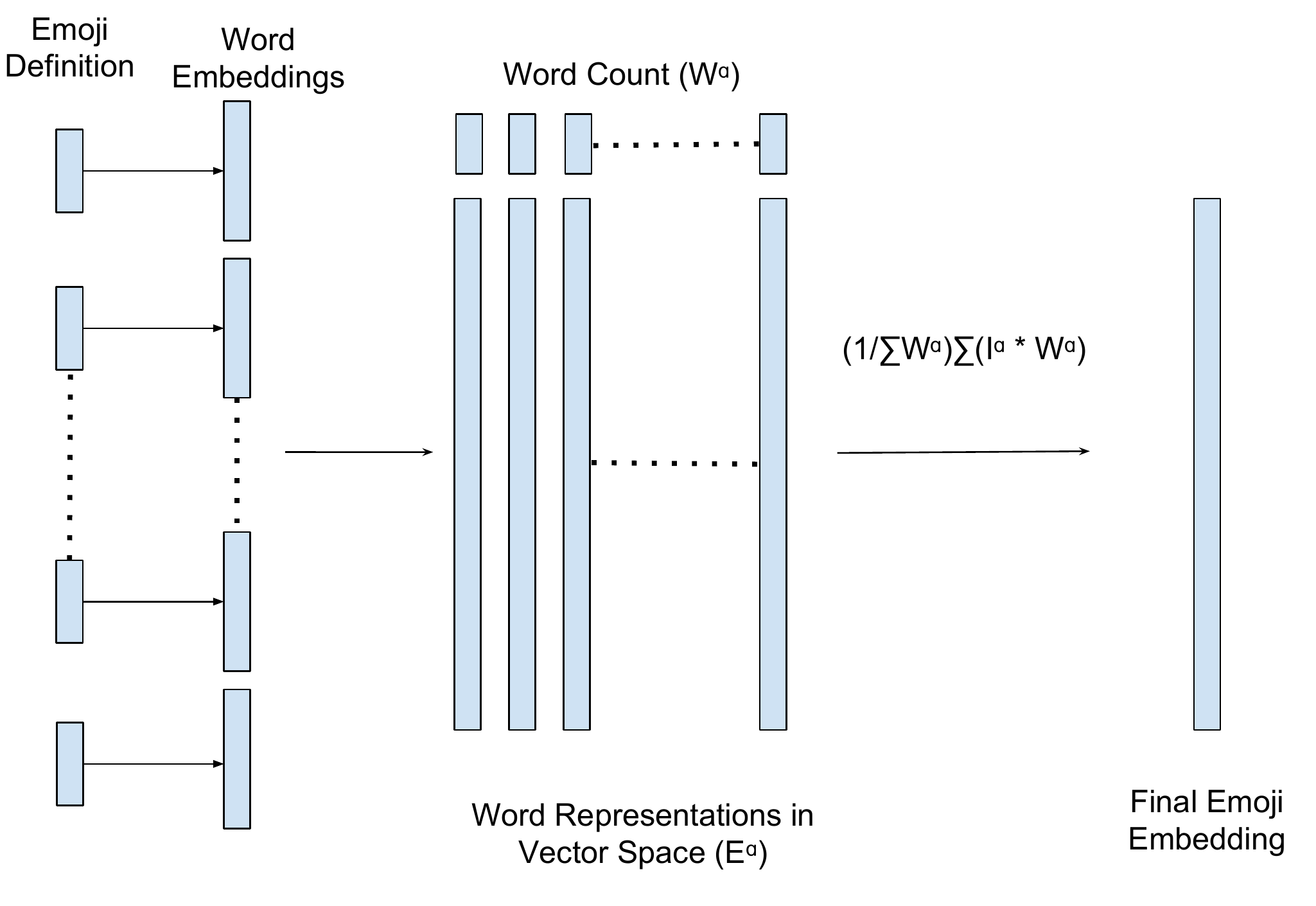} 
\vspace*{-7mm}
\caption{Generation of Emoji Embeddings}.
\label{fig4}
\end{figure}

We extracted the captions corresponding to each image of MSCOCO validation dataset and trained a FastText \cite{joulin2016FastText} word embedding model to learn word representations in finite-dimensional vector space. We also used the pre-trained Google-News\footnote{\url{https://bit.ly/1R9Wsqr}} word embedding model trained using word2vec \cite{goldberg2014word2vec} and pre-trained FastText model trained on Wikipedia corpus \cite{mikolov2018advances} to evaluate our model. We make use of Emojinet which gathers knowledge concepts of 2389 emojis. Specifically, EmojiNet provides a set of words (also called as senses), its POS tag and its sense definitions. It maps 12,904 sense definitions to 2,389 emojis. We learn the emoji representations from these external knowledge concepts using the approach discussed by Wijeratne et al. \cite{wijeratne2017semantics}. They replaced the word vectors of all words in the emoji definition and formed a 300-dimensional vector performing vector average. Also, the vector mean (or average) adjusts for word embedding bias that could take place due to certain emoji definitions having considerably more words than others has been noted by Kenter et al. \cite{kenter2016siamese}. \hyperref[sec:bagof]{Figure 4} illustrates the emoji embeddings model used to learn emoji representations from emoji knowledge concepts. We use the standard pre-processing techniques which include removing stop-words, removing articles and lemmatizing each word of emoji sense definitions and get another set of knowledge concepts which are referred as processed emoji sense definitions in later sections of the paper. We learn three types of emoji embeddings Emoji\_Embeddings\_Senses, and Emoji\_Embeddings\_Descriptions, and Emoji\_Embeddings\_Processed\_Descriptions using emoji senses, emoji sense definitions, and processed emoji sense definitions respectively.  The model of approach is evaluated using these emoji embeddings as knowledge concepts. 

$Emoji\_Embeddings\_Senses$ : The emoji sense forms is a list of different senses of what emoji mean in different contexts. EmojiNet lists ``love", ``face", ``beloved", ``dear", ``adorable" etc to be sense forms for the emoji ``face blowing a kiss" ( \includegraphics[height=1em]{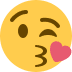} ). The emoji embedding, in this case, is the vector average of word embeddings as described in \hyperref[sec:bagof]{Figure 4}. The equation corresponding to calculation of emoji embedding, in this case, is given below where $\vec V_i$ represent word embedding of word $W_i$, n represents the number of distinct sense forms : 

\begin{equation}
    Emoji\_Embeddings\_Senses = \frac{\sum \vec V_i}{n}
\end{equation}

$Emoji\_Embeddings\_Descriptions$: The emoji definitions are textual descriptions which explain the context of use of an emoji. EmojiNet lists ``Love is a variety of different feelings, states, and attitudes that ranges from interpersonal affection to pleasure.", ``An intense feeling of affection and care towards another person." as some of the descriptions for the emoji ``face blowing a kiss"( \includegraphics[height=1em]{967.png} ). The equation corresponding to the emoji embedding is given below where $C_i$ represents the number of occurrences of word $W_i$, $\vec V_i$ is the word embedding of $W_i$:

\begin{equation}
    Emoji\_Embedding =  \frac{\sum \vec V_i \cdot C_i}{\sum C_i}
\end{equation}
\vspace*{-7mm}

\subsection{Model Architecture}

We use a pre-trained Resnet-152 \cite{He2015} image classifier, a $152$ layered Residual Network for image classification. Resnet-152 predicts the probability that an image belongs to a particular class. We replace the class label with its corresponding word embeddings learned using the word embedding models as discussed earlier and we call this the class embedding \cite{akata2016label}. Many researchers \cite{barbieri2018multimodal,Galanopoulos2015} have worked on combining textual and visual features for improving accuracies of multimedia tasks in the fields of NLP and Image processing. Using the probabilities predicted by Resnet-152 classifier we calculate the image embedding which combines the textual features and embeds the image to the similar embedding space as words \cite{karpathy2015deep}. This image embedding helps us to visualize an image in the same vector space as words.  Let $W_i$, $\vec C_i$ denote class label and word embedding of the class label respectively, $P_i$ denote probability associated with this class. We compute the image embedding using

\vspace*{-3mm}
\begin{equation}
    Image\_Embedding = \sum \vec{C_i}*P_i
\end{equation}

\begin{figure}
\centering
\includegraphics[width=1.0\linewidth]{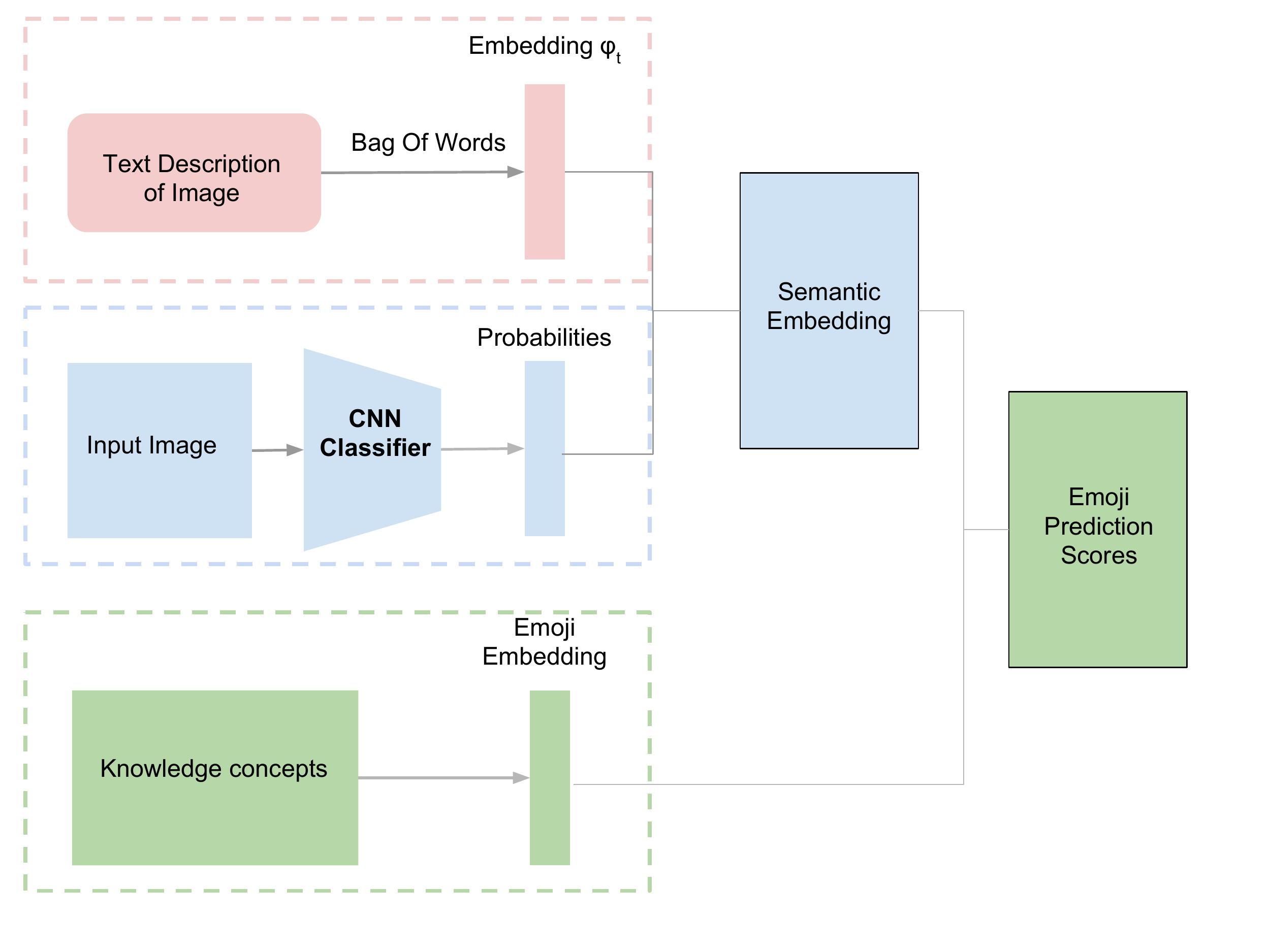} 
\caption{Model Architecture}.
\label{fig5}
\end{figure}

\begin{figure}
\label{sec:imageandemoji}
\centering
\includegraphics[width=1.0\linewidth]{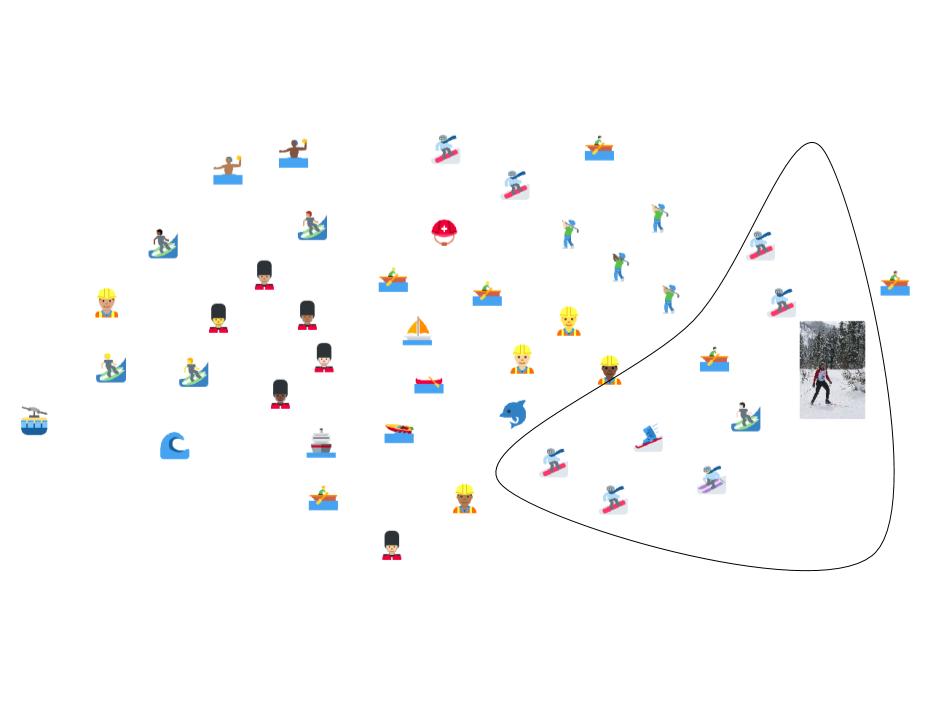} 
\vspace*{-20mm}
\caption{Embedding (Image+Textual description) and Emojis to similar vector space which helps us to calculate similarity between entities. The curve groups the emojis which are closer and similar to the context of the image}.
\label{fig6}
\end{figure}

We hypothesize that this image embedding helps us compute the context of the image using the word representation of the image classes, thus further helping us to predict an emoji in the context of the image. The image caption helps us understand the context of use of the image has been noted by Barbieri et al. \cite{barbieri2018multimodal}. Hence we use image caption as an additional feature to learn image representation. We use the same bag-of-words model illustrated in \hyperref[sec:bagof]{Figure 4} (approach used to calculate emoji representations) to calculate the representation of the image caption in low dimensional vector space. We use vector addition operation to combine the caption embedding and the image embedding as both representations are embedded in the similar vector space \cite{ha2018characterizing}, we term the combined embedding as image embedding in further sections. Consider \hyperref[sec:imageandemoji]{Figure 6} where we represent the emojis and the image embedding calculated using our approach on the same vector space. We calculated the 300-dimensional emoji representations using knowledge concepts from EmojiNet and the image embedding using our approach, and we use pre-trained FastText word embedding model on Wiki Corpus \footnote{\url{https://bit.ly/2FMTB4N}} \cite{mikolov2018advances} for word representations. Since one cannot visualize 300-dimensional vectors, we use the tSNE visualization \cite{maaten2008visualizing} to project the 300-dimensional emoji representations to two-dimensional vector space. We observed that emojis which are most similar to the context of the image are closer to the image compared to other emojis. Hence we could justify that this image embedding helps us in this emoji recommendation task. Thus this further adds a strong argument to combine both visual and textual features for emoji scoring. Each emoji is scored according to the similarity between image embedding (visualizes the context of the image) and the emoji embeddings (visualizes the context of use of an emoji), and we use cosine similarity as the distance measure. We term this task as emoji scoring in further sections of the paper. We evaluate our model using the two emoji embeddings as knowledge concepts and report our results on some images extracted from MSCOCO in \hyperref[sec:resultstable]{Table 4}. We observe that the emojis recommended by our model are in context with the image.

\section{Experiments}
\label{sec:experiments}

\subsection{Twitter Dataset}

We use our emoji recommendation model to predict the emoji which can be used in the context of the image by emoji scoring and calculate the number of tweets where the actual emoji label used in the tweet is the emoji predicted by our model. \hyperref[20emojis]{Table 1} reports the percentage of tweets in which the emoji label is the emoji predicted by our model. We considered image embedding as the visual feature (V) and the combination of image embedding and textual embedding as the combined visual and textual feature (V + T) to evaluate our model. We evaluated it on three different word embedding models namely Google News word embedding model, FastText trained on MSCOCO Descriptions, FastText trained on entire Wikipedia corpus \cite{mikolov2018advances} and using four external knowledge concepts namely Emoji Names, Emoji Sense Forms, Emoji Sense Definitions and Processed Emoji Sense Definitions. All the results in this paper report the number of tweets where the emoji used in the tweet is the emoji recommended by our model.

\begin{table}[H]
\centering
\caption{Percentage of Tweets in which the emoji used in the tweet is the emoji recommended by our model (These accuracies are observed if we considered the top 20 most frequently occurring emojis in \hyperref[sec:emojicount]{Twitter Dataset} for emoji scoring) (V - Visual features, V + T - combined Visual and textual features)}
\label{20emojis}

\begin{tabular}{|c|c|c|c|c|}
\hline
\multirow{3}{*}{Word Embedding Model} & \multicolumn{4}{c|}{Knowledge Concepts}                                                                                                                             \\ \cline{2-5} 
                                      & \multicolumn{2}{c|}{\begin{tabular}[c]{@{}c@{}}Emoji\\ Names\end{tabular}} & \multicolumn{2}{c|}{\begin{tabular}[c]{@{}c@{}}Emoji\\ Sense Definitions\end{tabular}} \\ \cline{2-5} 
                                      & V                                    & V+T                                 & V                                          & V+T                                       \\ \hline
Google News Word Embeddings           & 29.9\%                               & 31.2\%                              & 40.1\%                                     & 40.9\%                                    \\ \hline
FastText trained on MSCOCO            & 31.8\%                               & 32.9\%                              & 41.8\%                                     & 43.9\%                                    \\ \hline
FastText trained on Wiki Corpus       & \textbf{32.3\%}                               & \textbf{34.8\%}                              & \textbf{42.3\% }                                    & \textbf{45.1\%}                                    \\ \hline
\end{tabular}
\end{table}

\subsection{User Study}

As discussed earlier in \hyperref[sec:dataset]{Section 3}, each image in the MSCOCO dataset is annotated by three annotators. We observed a high accuracy score for MSCOCO dataset as the descriptions of each image listed in MSCOCO explains the context of the image more effectively as compared to the user descriptions on social media platforms, this helped our model to effectively capture the context of the image using the textual descriptions. \hyperref[userstudy]{Table 2} reports the number of images where the emoji label selected most times by three annotators is the emoji recommended by our model. We also report the number of images where the emoji label is one of the top-3 emojis predicted by our model. We used Cohen's kappa coefficient $(\kappa)$ to measure the inter-rater agreement to be 0.664 which is a good agreement value ($ 0.6 < \kappa < 0.8 $).

\begin{table}[H]
\caption{Number of Images where user annotated emoji label belongs to set of emoji recommendations by our model}
\label{userstudy}
\begin{center}
\begin{tabular}{|c|c|c|c|c|}
\hline
\multirow{2}{*}{}&\multicolumn{4}{c|}{Knowledge Concepts}\\ \cline{2-5}
&\begin{tabular}[c]{@{}c@{}}Emoji\\  Names\end{tabular} & \begin{tabular}[c]{@{}c@{}}Emoji \\ Senses\end{tabular} & \begin{tabular}[c]{@{}c@{}}Emoji \\ Definitions\end{tabular} & \begin{tabular}[c]{@{}c@{}}Processed \\ Emoji Definitions\end{tabular} \\ \hline
top-1 & 148                                                     & 311                                                     & 217                                                     & 356                                                                    \\ \hline
top-3 & 224                                                     & 386                                                     & 278                                                     & 426                                                                    \\ \hline
\end{tabular}
\label{tab1}
\end{center}
\end{table}


\section{Discussion}
\label{sec:discussion}

To further demonstrate the effectiveness of the proposed method, we compare it with the state-of-the-art image2emoji models for emoji prediction. \hyperref[sec:resultstable]{Table 4} summarizes the results obtained by our model and the image2emoji model \cite{cappallo2015image2emoji}. The third and fourth column report the top 5 emojis arranged according to their score when processed emoji sense definitions, emoji senses are used as external knowledge concepts respectively. The last column reports the predicted emojis using the image2emoji model. Consider the set of emojis predicted in the context of 4th image in \hyperref[sec:resultstable]{Table 4}, the emojis predicted using the image2emoji model does not closely relate to the context of the image, whereas the emojis predicted using processed emoji definitions as external knowledge concepts are more relevant in the context of the image. Further, it can be noted that the recommendations obtained by considering processed emoji sense definitions as external knowledge concepts are more relevant compared to other sets of predictions, this is due to the fact the processed emoji sense definitions explain the context of use of an emoji.

\begin{table*}
\centering
\caption{Percentage of Tweets in which the emoji used is one of the top-3 emojis recommended by our model\newline (These accuracies are observed if we considered the complete set of emojis in Twitter dataset (1089 emojis) for emoji scoring) (V - Visual features, V + T - combined Visual and textual features)}

\label{percentage}
\label{1089emojis}
\begin{tabular}{|c|c|c|c|c|c|c|c|c|}
\hline
\multirow{3}{*}{Word Embedding Model} & \multicolumn{8}{c|}{Emoji Knowledge Concepts} \\ \cline{2-9} 
                                      & \multicolumn{2}{c|}{Emoji Names} & \multicolumn{2}{c|}{Emoji Sense Forms} & \multicolumn{2}{c|}{Emoji Sense Definitions} & \multicolumn{2}{c|}{Processed Sense Definitions} \\ \cline{2-9} 
                                      & V & V + T & V & V + T & V & V + T & V & V + T \\ \hline
Google News Word Embeddings           & 4.08\% & 5.72\% & 14.76\% & \textbf{18.49\%} & 8.46\% & 12.03\% & 19.21\% & 21.23\% \\ \hline
FastText on MSCOCO Descriptions       & 4.71\% & 6.63\% & 15.34\% & 18.24\% & \textbf{9.46\%} & 12.89\% & \textbf{19.62\%} & 22.04\% \\ \hline
FastText on Wikipedia Corpus          & \textbf{5.45\%} & \textbf{7.02\%} & \textbf{15.61\%} & 17.89\% & 9.02\% & \textbf{13.43\%} & 19.45\% & \textbf{22.49\%} \\ \hline

\end{tabular}
\end{table*}

\hyperref[20emojis]{Table 1} and \hyperref[1089emojis]{Table 3} report the accuracies observed considering top 20 most frequent emojis and the complete set of 1089 emojis present in the twitter dataset and scored according to their relevance to the context of the image respectively. Francesco et al. \cite{barbieri2018multimodal} has reported that they have achieved a accuracy of 35.5\% when they considered top-20 most frequent emojis as labels for emoji prediction. Our model has outperformed and achieved a accuracy of 45.1\% if top-20 most frequently used emojis in Twitter dataset are considered as labels for emoji recommendation. We observed an accuracy of 22.49\% (our model could predict the correct emoji label in 6102 images out of 27136 images) when processed emoji sense definitions are considered as external knowledge concepts and 1089 emojis are used as labels for emoji scoring. Hence this further demonstrates the effectiveness of the proposed approach with a huge set of 1089 emoji labels. Also, it can be noted that in most cases FastText trained word embeddings on Wikipedia corpus have resulted in high accuracies as the fastText model is proved to result in high accuracies in most NLP tasks compared to other word embedding models \cite{mikolov2018advances}.


\section{Conclusion and Future Work}
\label{sec:conclusion}

In this paper, we introduced a knowledge-enabled emoji recommendation system which helps users select an emoji which talks better for an image or picture using domain knowledge from EmojiNet. Experimental results show that our results have outperformed the previous and current state-of-the-art results for the image2emoji models \cite{cappallo2015image2emoji,barbieri2018multimodal}. \hyperref[sec:resultstable]{Table 4} reports some of the exciting emoji recommendations using various knowledge concepts for emoji scoring. Accuracy of our model has been observed to be more if processed emoji sense definitions are used as external knowledge concepts for emoji scoring. We further plan to extend our work by introducing deep learning models for emoji predictions in the context of images. Venugopalan et al. (\cite{venugopalan2016improving})  have used linguistic knowledge from large text corpora to generate natural language descriptions of videos. Using this as a reference, we plan to extend our work in future by building models which can summarize a video to a sequence of meaningful emojis which convey the same visual content of a video, using existing domain knowledge for emojis.

\section*{Acknowledgement}
We are grateful to Swati Padhee, Sanjaya Wijeratne and Dr. Amit P. Sheth for thought-provoking discussions on the topic. We acknowledge support from the Indian Institute of Technology Kharagpur and Indian Institute of Technology Hyderabad. Any opinions, findings, and conclusions/recommendations expressed in this material are those of the author(s) and do not necessarily reflect the views of Indian Institute of Technology Kharagpur and Indian Institute of Technology Hyderabad.

\begin{table*}[ht]
\begin{center}
\caption{Top 5 Emojis Predicted in the context of an image using different emoji knowledge concepts}
\label{sec:resultstable}
\begin{tabular}{| >{\centering\arraybackslash}m{0.5in}|
>{\centering\arraybackslash}m{1in} | >{\centering\arraybackslash}m{1in} |
>{\centering\arraybackslash}m{1in} |
>{\centering\arraybackslash}m{1in} |
>{\centering\arraybackslash}m{1in} |}
\hline
 \textbf{S.No} & \textbf{Image} & \textbf{Text Description} & \textbf{Using Processed sense definition} & \textbf{Using emoji senses} & \textbf{Using emoji names} \\ 
 \hline
 1 &
 \includegraphics[height=0.7in,width=0.7in]{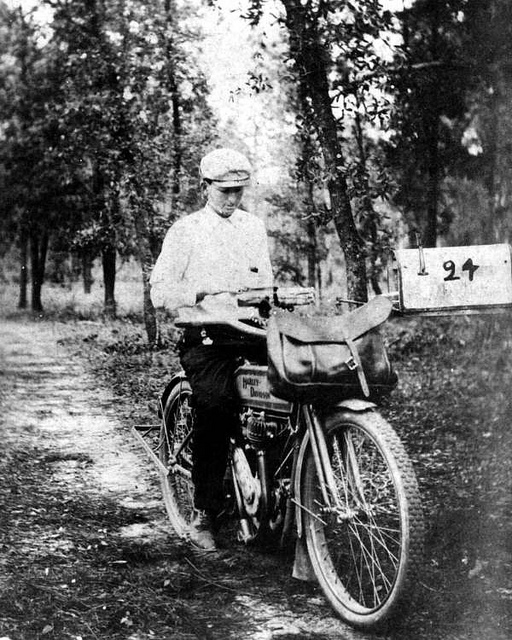} 
 & 
 A person looks down at something while sitting on a bike 
 & 
\includegraphics[height=1em]{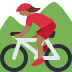}
 \includegraphics[height=1em]{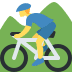}
 \includegraphics[height=1em]{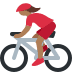}
 \includegraphics[height=1em]{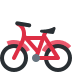}
 \includegraphics[height=1em]{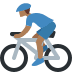}
 &
 \includegraphics[height=1em]{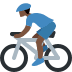}
 \includegraphics[height=1em]{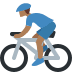}
 \includegraphics[height=1em]{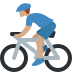}
 \includegraphics[height=1em]{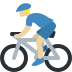}
 \includegraphics[height=1em]{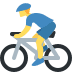}
 &
 \includegraphics[height=1em]{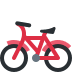}
 \includegraphics[height=1em]{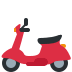}
 \includegraphics[height=1em]{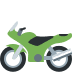}
 \includegraphics[height=1em]{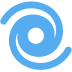}
 \includegraphics[height=1em]{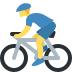}
 
 \\
 \hline
 2 &
 \includegraphics[height=0.7in,width=0.7in]{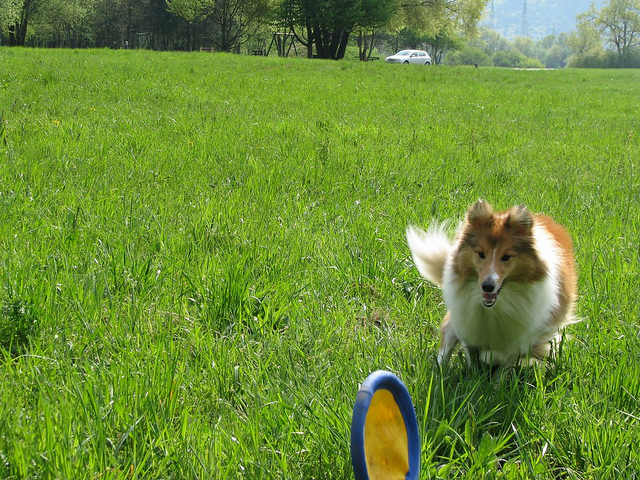} 
 & 
 The dog is playing with his toy in the grass 
 & 
 \includegraphics[height=1em]{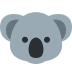}
 \includegraphics[height=1em]{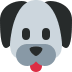}
 \includegraphics[height=1em]{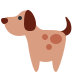}
 \includegraphics[height=1em]{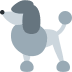}
 \includegraphics[height=1em]{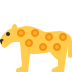}
 &
 \includegraphics[height=1em]{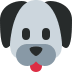}
 \includegraphics[height=1em]{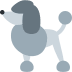}
 \includegraphics[height=1em]{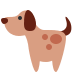}
 \includegraphics[height=1em]{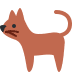}
 \includegraphics[height=1em]{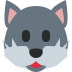}
 &
 \includegraphics[height=1em]{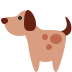}
 \includegraphics[height=1em]{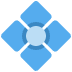}
 \includegraphics[height=1em]{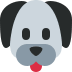}
 \includegraphics[height=1em]{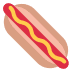}
 \includegraphics[height=1em]{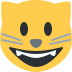}
 
 \\
 \hline
 3 &
 \includegraphics[height=0.7in,width=0.7in]{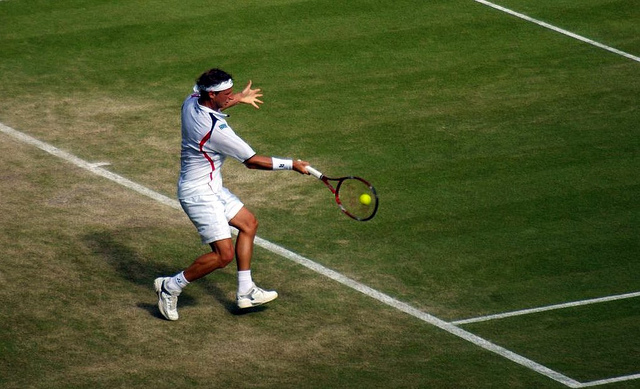} 
 & 
 A tennis player in action on the court 
 & 
 \includegraphics[height=1em]{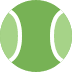}
 \includegraphics[height=1em]{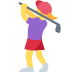}
 \includegraphics[height=1em]{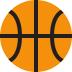}
 \includegraphics[height=1em]{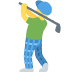}
 \includegraphics[height=1em]{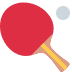}
 &
 \includegraphics[height=1em]{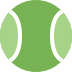}
 \includegraphics[height=1em]{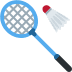}
 \includegraphics[height=1em]{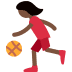}
 \includegraphics[height=1em]{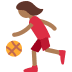}
 \includegraphics[height=1em]{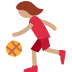}
 &
 \includegraphics[height=1em]{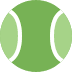}
 \includegraphics[height=1em]{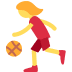}
 \includegraphics[height=1em]{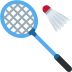}
 \includegraphics[height=1em]{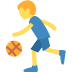}
 \includegraphics[height=1em]{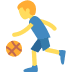}
 \\
 \hline
 4 &
 \includegraphics[height=0.7in,width=0.7in]{results/img_4/COCO_train2014_000000000605.jpg} 
 & 
 Cup of coffee with dessert items on a wooden grained table 
 & 
 \includegraphics[height=1em]{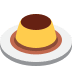}
 \includegraphics[height=1em]{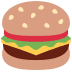}
 \includegraphics[height=1em]{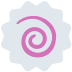}
 \includegraphics[height=1em]{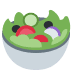}
 \includegraphics[height=1em]{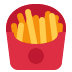}
 &
 \includegraphics[height=1em]{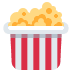}
 \includegraphics[height=1em]{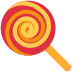}
 \includegraphics[height=1em]{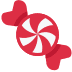}
 \includegraphics[height=1em]{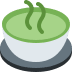}
 \includegraphics[height=1em]{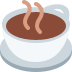}
 &
 \includegraphics[height=1em]{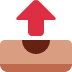}
 \includegraphics[height=1em]{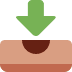}
 \includegraphics[height=1em]{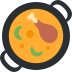}
 \includegraphics[height=1em]{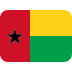}
 \includegraphics[height=1em]{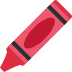}
 \\
 \hline
 
\end{tabular}
\end{center}
\end{table*}

\vspace*{-1.5mm}

\bibliographystyle{plain}

\end{document}